%% file: acl_latex.tex
\documentclass[11pt]{article}

\usepackage[preprint]{acl}

\usepackage{times}
\usepackage{latexsym}

\usepackage[T1]{fontenc}
\usepackage[utf8]{inputenc}

\usepackage{microtype}

\usepackage{inconsolata}

\usepackage{graphicx}
\usepackage{amsmath}
\usepackage{amssymb}
\usepackage{booktabs}
\usepackage{colortbl}
\usepackage{multirow}
\usepackage{array}
\usepackage{enumitem}
\usepackage{pifont}
\usepackage{url}
\usepackage{listings}
\newcommand{\benchfull}{Terminal-style Mathematical Modeling Bench}
\newcommand{\bench}{TB-MathModel}
\newcommand{\oeval}{\textsc{O-Eval}}

\newcommand{\roeval}{\textsc{Robust BO-Eval}}
\newcommand{\boeval}{\textsc{Multi-BO}}

\title{\benchfull{}:\\ The First Agentic and Verifiable Benchmark for Mathematical Modeling}

\author{
\bf Xiaojun Wu$^{\ast\ 1,2}$,
Cehao Yang$^{\ast\ 1,2}$,
Xuhui Jiang$^{1,3}$,  \\
\bf Chengjin Xu$^{1,3}$,
Jia Li$^{\dagger\ 2}$,
Jian Guo$^{\dagger\ 1}$
\\
\\
$^{1}$IDEA Research \\
$^{2}$The Hong Kong University of Science and Technology (Guangzhou) \\
$^{3}$DataArcTech Ltd.
}

\begin{document}
\maketitle
{
  \renewcommand{\thefootnote}%
    {\fnsymbol{footnote}}
  \footnotetext[1]{Equal Contribution}
  \footnotetext[2]{Corresponding Author}
}
\begin{abstract}
\input{contents/0.Abstract}
\end{abstract}

\input{contents/1.Introduction}
\input{contents/2.Related_work}
\input{contents/3.Methodology}

\input{contents/4.Experiment}

\input{contents/5.Conclusion}

\input{contents/Acknowledgements}

\clearpage
\input{contents/Limitations}
\input{contents/LLM_usage}

\clearpage
\bibliography{custom}

\clearpage
\appendix
\input{contents/Appendix/0.Appendix_overview}

\end{document}

%% file: contents/0.Abstract.tex
Mathematical modeling turns an open-ended real-world problem into assumptions, a model, a solution, and a defended recommendation, and its answers matter only if they are \emph{both} close to a credible reference \emph{and} physically feasible; existing benchmarks either score solver fragments or grade reports with an LLM judge, leaving those numbers unchecked. We introduce \bench{} (the \benchfull{}), the first benchmark to make complete contest-style modeling problems agentic, verifiable, and terminal at once, over 18 real CUMCM and MCM problems (2023--2025). Each task ships the verbatim statement, official attachments, and figures in a containerized terminal; the agent returns one JSON artifact, scored on the numeric core of the final question against a reproduction of the Outstanding-award paper. Two design choices distinguish it: \emph{direction-aware, one-sided} scoring, which never penalises beating the anchor, and \emph{two-layer feasibility verification}---a programmatic gate over six decidable invariant forms plus 624 Lean~4 obligations---with a hard gate that zeroes infeasible answers. Each task is scored against a panel of human outstanding papers rather than one, so a model must be close to many accepted answers; across seven multimodal entries (three seeds each) the strongest reach a multi-reference \oeval{} of 64.58\%, while the \$2.54 open model holds 59.81\% at $1/100$ the cost. \bench{} turns the numeric core of modeling into a reproducible, machine-checkable signal for measuring long-horizon modeling competence and for collecting agent trajectories. The source code is available at \url{https://github.com/wxj630/terminal-bench-math-modeling}.

%% file: contents/1.Introduction.tex
\section{Introduction}

Mathematical modeling contests such as CUMCM in China and MCM/ICM internationally ask teams to convert a deliberately under-specified real-world problem into a defensible quantitative answer within days. The deliverable is not a single number but a chain: assumptions $\rightarrow$ model $\rightarrow$ solution $\rightarrow$ sensitivity $\rightarrow$ recommendation. Two properties make it an attractive target for agent evaluation: it is \emph{quantitative}, because the final question ends in numbers that can be checked, and it is \emph{multimodal}, because statements reference figures and attachments are spreadsheets and CSVs. It also differs from solving a well-posed problem in \emph{where the hard work begins}---the model itself must be built, and the same situation admits time-series, optimization, dynamical, graph, or multi-criteria formulations, any of which can be defensible if its assumptions are explicit and its computation credible~\cite{Giordano_Horton_Fox_2014,meerschaert2013mathematical}. This is what makes modeling a demanding test of long-horizon agents: hold the statement and attachments in context, choose a formulation, write and run code, handle files and units, and report a self-consistent answer.

Existing benchmarks cover adjacent but distinct ground, and none closes the loop we care about. Code, terminal, and assistant benchmarks score repository edits, command-line competence, and tool use~\cite{swebench,terminalbench,gaia,taubench}; closed-form mathematics benchmarks score deduction rather than modeling~\cite{hendrycks2021MATH,cobbe2021gsm8k,frontiermath}; operations-research datasets parse word problems into solver-ready formulations~\cite{nl4opt,optimus,optibench}. The closest prior work on modeling itself falls into two camps, neither simultaneously agentic, verifiable, terminal, and multimodal. One stays \emph{solver-facing} and narrow, scoring a single formal artifact~\cite{Mamo}. The other grades the final \emph{report} with an LLM judge~\cite{ModelingAgent,MMAgent}, rewarding fluent prose without establishing that the reported numbers are correct or even admissible. Table~\ref{tab:compare} makes the gap concrete; \bench{} is the first to combine all of these properties.

\begin{table*}[t]
\centering
\small
\setlength{\tabcolsep}{6pt}
\caption{The closest mathematical-modeling benchmarks. Only \bench{} combines a complete contest problem, an agentic terminal run, an executable verifier, physical-feasibility checking, and a machine-checkable proof, on a multimodal task (``Judge'': LLM grades a report; ``Solver'': solver checks one formal artifact).}
\label{tab:compare}
\begin{tabular}{@{}lrrlcccc@{}}
\toprule
Benchmark & \#Tasks & \#Fig. & Evaluation signal & Agentic & Physical & Lean & Multimodal \\
\midrule
Mamo~\cite{Mamo}                   & 1209 & -- & Solver check        & \ding{55} & \ding{55} & \ding{55} & \ding{55} \\
ModelingBench~\cite{ModelingAgent}& 68   & -- & LLM judge           & \ding{55} & \ding{55} & \ding{55} & \ding{55} \\
MM-Bench~\cite{MMAgent}            & 111  & -- & LLM judge           & \ding{55} & \ding{55} & \ding{55} & \ding{55} \\
\bench{} (ours)                    & 18   & 21 & Executable verifier & \ding{51} & \ding{51} & \ding{51} & \ding{51} \\
\bottomrule
\end{tabular}
\end{table*}

Closing the gap raises three problems that shape the design. Different teams make different assumptions, so there is no single ground truth: rather than promote one paper to truth, \bench{} \emph{asks every human expert}, scoring each task against a panel of outstanding solutions so that a model must be close to many accepted answers, not match one. Anchoring creates asymmetric failure---a model that does \emph{better} than the paper can score worse than one that copies it---so \bench{} makes each metric's direction explicit and scores one-sidedly. And a wrong-but-plausible answer is easy to produce (numbers that look reasonable yet violate the problem's physics), which anchor comparison cannot catch, so \bench{} adds an independent feasibility layer.

We present \bench{} (\benchfull{}), to our knowledge the \emph{first} benchmark to turn \emph{complete contest-style mathematical modeling problems} into an \emph{agentic, verifiable, terminal} evaluation. It contributes:
\begin{enumerate}[leftmargin=1.5em,itemsep=2pt]
\item \textbf{The first benchmark of its kind, and the 18-task dataset behind it.} No prior benchmark turns complete contest-style modeling problems into a containerized terminal task with an executable verifier. \bench{} covers CUMCM and MCM 2023--2025 (problems A/B/C each year), with verbatim statements, official attachments, 21 figures cropped from the PDFs, and one JSON artifact per task.
\item \textbf{Direction-aware, one-sided scoring with auditable edge-case rules.} Per-metric semantic directions (\texttt{higher/lower\_is\_better}, \texttt{target\_value}, \texttt{exact\_value}); two reported metrics (\oeval{}, \roeval{}); explicit rules for degenerate bands and unpinned reference units; and contract tests that lock the two defining properties down.
\item \textbf{A two-layer feasibility verification design.} A programmatic gate covering 18/18 tasks through six decidable invariant forms, and a Lean~4 certificate layer of 624 kernel-checked obligations; the two layers decide every feasibility question a structured artifact can settle, and state exactly where an independent replay takes over.
\item \textbf{An empirical study of seven multimodal model/harness entries}, each averaged over three seeds, with per-task behaviour, a cost--robustness analysis, and three case studies; we also document the scoring pitfalls we found and fixed and quantify which move the leaderboard.
\end{enumerate}

Against the human panel, the two newest flagships lead (GPT-6 Astra + codex at \oeval{} $64.58 \pm 2.93$ and \boeval{} $1.132$, Fable 5.1 + claude-code at $63.11 \pm 1.33$ and $1.081$, means over three seeds), while the \$2.54 DeepSeek~V4.1~Flash + opencode holds $59.81 \pm 1.25$ at $1/100$ the cost; varying only the harness moves \oeval{} by at most 0.07~pp, so the board measures the model rather than the harness. Per-seed deviations reach $2.9$~pp, so adjacent rows are not significantly separated. One case study inverts the usual story: on dragon-dance the models were right and our anchor was wrong---the sharpest argument for independently checkable feasibility.

\begin{figure*}[t]
\centering
\includegraphics[width=\textwidth]{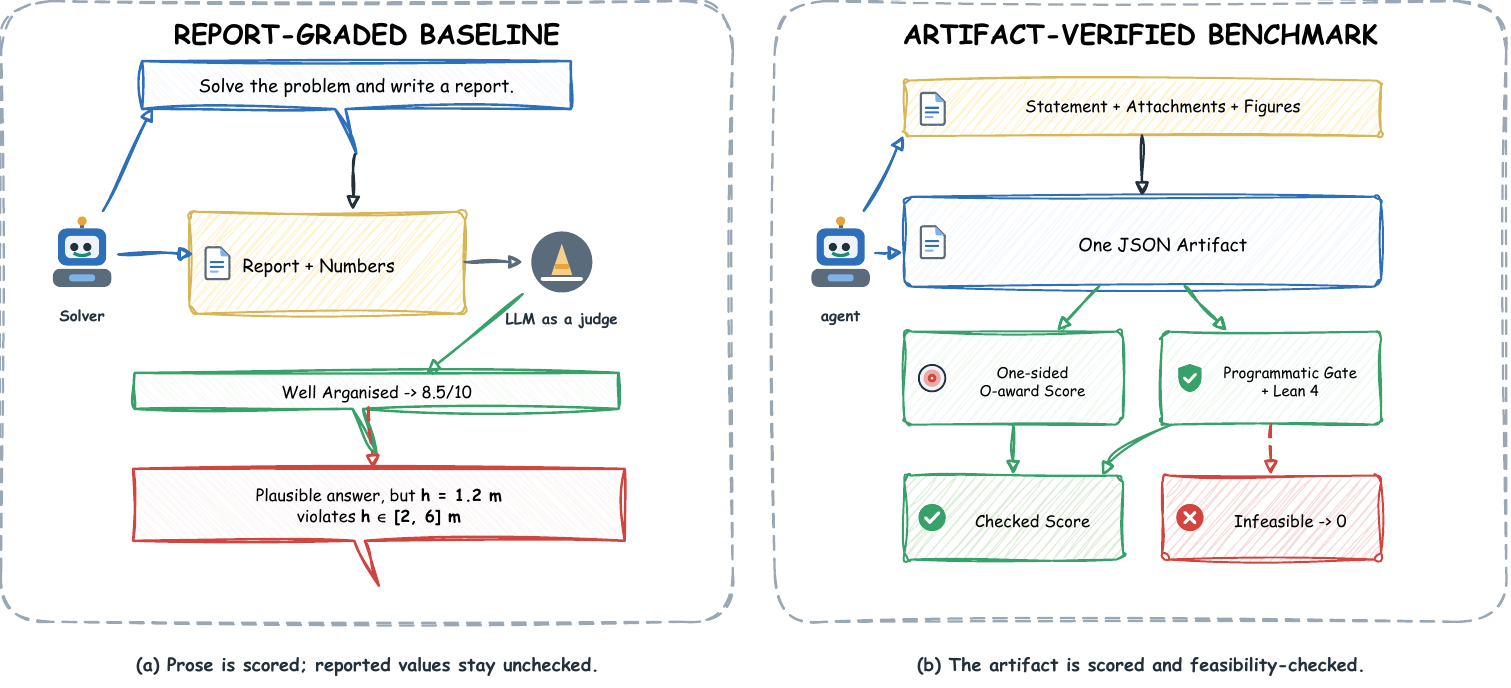}
\caption{Why modeling is hard to evaluate, and what \bench{} changes. (a)~In the status quo an LLM judge grades a report on organisation while the numbers are never checked. (b)~\bench{} runs the problem, attachments, and figures in a terminal container, requires one structured artifact, scores it one-sidedly against an O-award reproduction, and verifies feasibility with a gate and Lean certificates---so a plausible-but-infeasible answer is caught.}
\label{fig:motivation}
\end{figure*}

%% file: contents/2.Related_work.tex
\section{Related Work}

\paragraph{Agent and assistant benchmarks.}
Software-engineering and terminal agents are measured by repository-level tasks (SWE-bench~\cite{swebench}) and end-to-end command-line tasks (Terminal-Bench~\cite{terminalbench}); assistants by tool-use suites such as GAIA~\cite{gaia} and $\tau$-bench~\cite{taubench}. Machine-learning and data-science agents are measured on Kaggle-style pipelines~\cite{mlebench,paperbench,DS-Agent,zhang2024benchmarking,gu2024blade,hu2024infiagent,li2024autokaggle}. These capture tool use and long-horizon execution in a container---the harness we build on---but score a leaderboard position or a rubric over a deliverable, never a modeling artifact, and attach no feasibility contract.

\paragraph{Mathematical and optimization reasoning.}
Closed-form mathematics benchmarks measure deduction, not modeling~\cite{hendrycks2021MATH,cobbe2021gsm8k,frontiermath,minif2f}. Operations-research systems parse word problems into solver-ready formulations and evaluate optimality or feasibility of a \emph{given} formulation~\cite{nl4opt,xiao2024chainofexperts,optimus,optibench,Huang2025orlm,mostajabdaveh2025evaluating,lima2025trustworthy,xiao2025survey}. They supply a strong \emph{execution-validity} precedent---a solver certifies a solution is feasible or optimal \emph{for a stated model}---but cannot certify that the model captures the right abstraction, and none requires assumptions, sensitivity analysis, or a defended recommendation.

\paragraph{Mathematical-modeling benchmarks and agents.}
The closest work forms a continuum, and none of it is simultaneously agentic, verifiable, and terminal. Mamo~\cite{Mamo} is \emph{solver-grounded}: models emit solver-consumable artifacts and the solver is a falsifiable checker, giving a precise but narrow signal on formal fragments rather than whole contest problems. ModelingAgent and its ModelingBench~\cite{ModelingAgent} are \emph{report-level}: open-ended problems graded by an expert-style LLM judge---ecologically rich but judge-calibration dependent and non-reproducible. MM-Agent and MM-Bench~\cite{MMAgent} add a hierarchical modeling library, yet their primary evaluation remains a report rubric. \bench{} keeps the full contest statement and the agentic workflow, pins evaluation to a \emph{structured artifact} scored by an executable verifier, and adds a feasibility layer neither camp provides.

\paragraph{Verification in evaluation.}
Theorem provers such as Lean~4~\cite{lean4} with Mathlib~\cite{mathlib} have certified competition and research mathematics, and RL over formal proof has produced strong provers~\cite{alphaproof,deepseekprover}. Using a prover not to \emph{find} answers but to \emph{audit a benchmark's own feasibility decisions} is, to our knowledge, new: our Lean layer re-proves the programmatic gate's invariants, so an assertion becomes a machine-checkable fact~\cite{suttonverification}, at integer-and-rational arithmetic cost the kernel decides without the full library.

\paragraph{Multimodal agents and reproduction-anchored scoring.}
Tool-using agents increasingly externalize their state for reuse~\cite{DBLP:conf/iclr/YaoZYDSN023,DBLP:conf/nips/Wei0SBIXCLZ22,schick2023toolformer,zhou2026externalization}. \bench{} makes visual evidence load-bearing---figures are cropped from the official PDFs---so harnesses with and without an image channel can be compared on identical tasks. Anchoring a numerical benchmark on a published solution is standard practice; we show it creates three under-reported hazards (direction asymmetry, degenerate scoring bands, unpinned units) and give auditable rules and tests for each.

%% file: contents/3.Methodology.tex
\section{The \bench{} Benchmark}
\label{sec:benchmark}
\begin{figure*}[t]
\centering
\includegraphics[width=\textwidth]{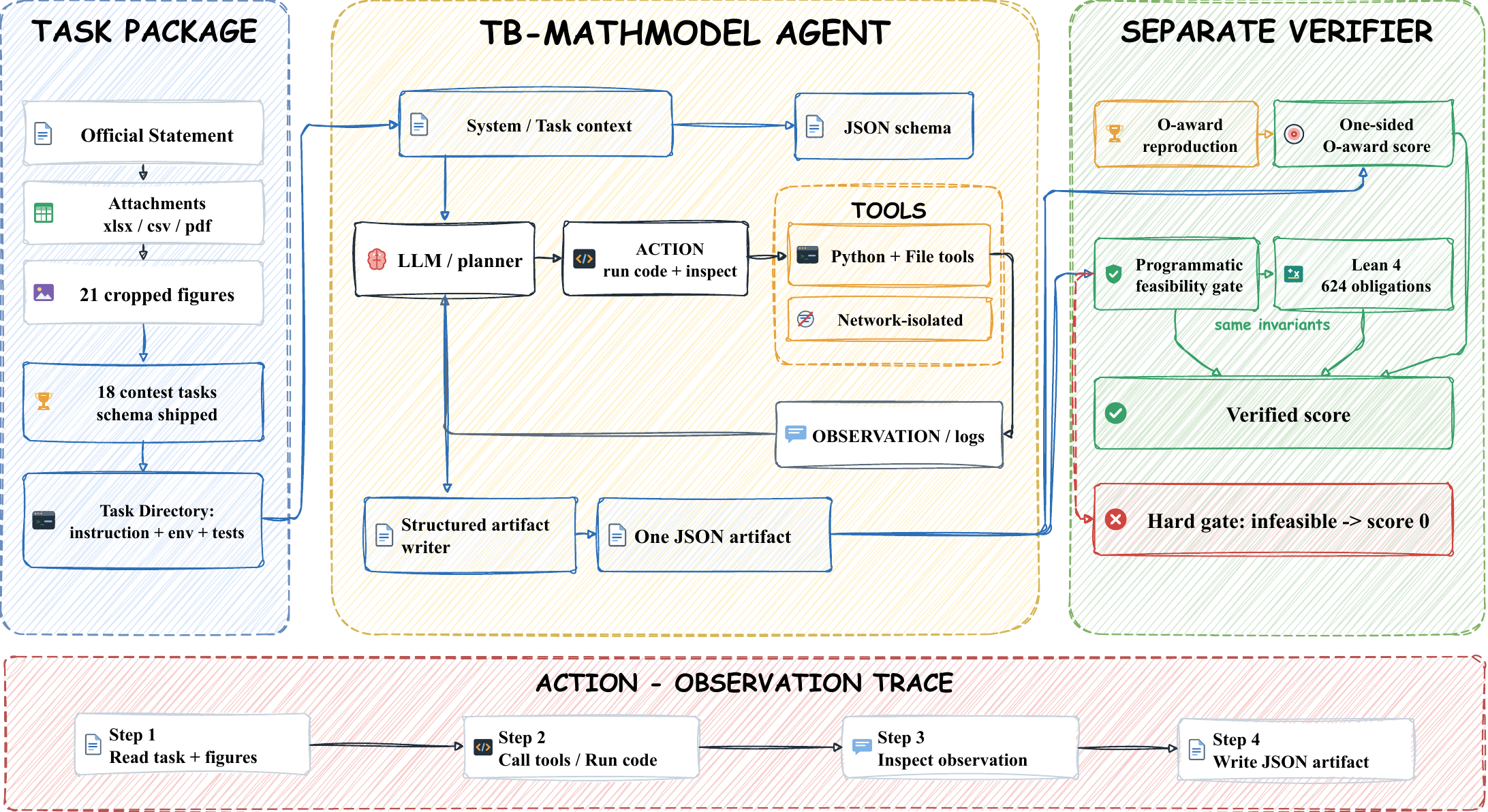}
\caption{The \bench{} construction and evaluation pipeline. \emph{Build}: official statements, attachments, and PDF figures are compiled into a task directory. \emph{Run}: the agent works inside a network-isolated terminal container and writes one JSON artifact. \emph{Evaluate}: the artifact is scored by a direction-aware, one-sided panel against an O-award reproduction, and independently checked by a programmatic feasibility gate plus a Lean~4 certificate layer; a failed gate zeroes the task.}
\label{fig:pipeline}
\end{figure*}

\bench{} (the \benchfull{}) turns each contest problem into a self-contained terminal task. Figure~\ref{fig:pipeline} summarises the construction pipeline. A build script assembles the official contest material into four parts: (i) an instruction file with the verbatim statement and its figures; (ii) an agent container holding the mirrored attachments and an output schema; (iii) a separate verifier container holding the scoring panel and an executable feasibility check; and (iv) an oracle that reproduces the Outstanding-award (O-award) paper. The agent is dropped into the terminal with this material and must return a single structured artifact. Everything downstream of the artifact---scoring and feasibility---is machine-executable.

\paragraph{Task selection and sources.}
\bench{} comprises \textbf{18 tasks}: the A, B, and C problems of CUMCM and MCM for each of 2023--2025. We restrict to the last three years on purpose: older problems predate capable LLMs and were calibrated to software-assistant graders, so their task mix and review standard differ from the post-LLM regime, and recent editions are harder and more discriminating. The two contests complement each other---CUMCM is engineering/optimization-heavy and Chinese, MCM is modeling-flavored and English---and taking all three problems of each year covers the standard ``A = physical/engineering, B = data/optimization, C = data-science'' split without cherry-picking. Statements and attachments are mirrored verbatim; a statement's figures are cropped from the official PDF (artwork plus its caption) and delivered inline and at stable container paths. \textbf{21 unique figures} across 6 tasks are \emph{load-bearing}: the statements reference them for geometry not restated in text (e.g., the dragon-dance U-turn curve, which fixes the final head speed), so a harness without an image channel cannot recover it---hence the multimodal evaluation. \bench{} is a pure evaluation set, with no training split or difficulty tiering (Appendix~\ref{app:dataset}).

\paragraph{The artifact contract and oracle anchors.}
The scored object is one JSON file per task, whose schema (field names and types only) is shipped to the agent; extra CSV/XLSX/PNG artifacts are not scored. A schema-constrained JSON is what makes an automated score possible, at the cost of mapping a free-form effort onto named numeric fields---a tension we revisit in Section~\ref{sec:analysis}. Each task is anchored to one O-award paper, re-implemented as a script that reads the attachments and emits the same JSON; the reproduction is the normalisation endpoint (\oeval{} $=100\%$) and sets the per-metric panel, coverage \textbf{18/18}. Provenance is a first-class field---paper id, PDF/OCR source, and reproduction scope---so any anchor can be re-derived (Appendix~\ref{app:baselines}); this is what let us catch and correct an infeasible dragon-dance anchor before it reached the board.

\subsection{Scoring: direction-aware, one-sided, auditable}
\label{sec:scoring}
A naive distance-to-anchor metric silently rewards copying and punishes improvement, so \bench{} scores each task through a per-metric panel with an explicit semantic direction. Each task ships a panel of 2--6 scored metrics (60 in total), each with a path into the artifact, an anchor value $O$, a direction, and a weight. A metric is \texttt{higher\_is\_better}, \texttt{lower\_is\_better}, \texttt{target\_value} (closest to $O$ wins), or \texttt{exact\_value}, set from the modeling semantics of the quantity rather than the sign of its value. Directional metrics use an asymmetric, \emph{one-sided} reward: meeting or beating the anchor never scores below $1.0$, so improving on the paper is never punished (full formulas in Appendix~\ref{app:scoring}).

Three failure modes recur often enough to deserve named rules. \textbf{(i) Degenerate bands:} a band narrower than 2\% of the value magnitude cannot discriminate, so we widen it to 5\%---this changed one cell (stair-wear, $72.67\!\to\!73.52$). \textbf{(ii) Non-comparable units:} when an anchor is an arbitrary raw index with no schema-pinned unit (a host-fish index of 1080), a model reporting the same quantity normalised to $[0,1]$ is not wrong, so such metrics are dropped for every model and the oracle by a configuration-only rule. \textbf{(iii) Cardinality mismatch:} surplus items are excluded rather than counted missing. Out-of-domain values (a correlation outside $[-1,1]$, a negative count) score zero, with the reason recorded.

Let $R$, $F$, $O$ be the weighted panel raw of a submission, of the DeepSeek~V4~Flash baseline, and of the oracle. We report \oeval{} $=\mathrm{clamp}(R/O,0,1)$ averaged over tasks (the headline absolute score) and a gap-compensated \roeval{} $=\mathrm{clip}((R-F)/(O-F),\pm1)$ when the oracle--baseline gap exceeds $0.10$, and the clipped raw gain $R-F$ otherwise, with $\roeval=-1$ for a missing artifact (full formulas in Appendix~\ref{app:scoring}). Two properties are locked by contract tests: \textbf{oracle self-consistency} (every reproduction scores exactly $1.0$) and \textbf{one-sidedness} (at or above the anchor is never a failure). A failed feasibility gate sets a task's \oeval{} to $0$ and \roeval{} to $-1$.

\subsection{Multi-references evaluation}
\label{sec:multi-ref}

Modeling has no single correct answer, so a benchmark that normalises against one paper silently converts one team's choices into ground truth. \bench{} instead asks every human expert. For each task we extracted the final-question values reported by \emph{all} available Outstanding-award papers for that problem, giving \textbf{99 panel papers, 94 of which yield the scored quantity}. Coverage is \textbf{18/18 tasks} and \textbf{all available outstanding papers are used}---1 to 18 per task (median 4), since the number of O-award reports differs by problem. Each paper is one vote.

Let $M(t)$ be the scored metrics of task $t$, $P(t)$ its paper panel, and $S_p \subseteq M(t)$ the fields paper $p$ actually reports. Writing $R(m,S)$ for the weighted panel score of submission $m$ restricted to $S$ (weights renormalised over $S$), the task score is the panel mean
\begin{equation}
\oeval(m,t) \;=\; \frac{1}{\lvert P_{\neq}(t)\rvert}
\sum_{p \in P_{\neq}(t)} R(m, S_p),
\end{equation}
where $P_{\neq}(t)=\{p \in P(t): S_p \neq \emptyset\}$. Each paper thus contributes exactly ``how close is the model to \emph{this} human answer, on the fields that human reported''. Because $R(p,S_p)=1$, every paper agrees with itself and the reproduction reproduces the panel. Equal weighting keeps one human solution to one vote, and a paper that reports only a single field acts as a single-field check rather than being discarded---the honest treatment of an incomplete report.

A paper is never silently dropped. When it cannot supply the scored quantity it carries a machine-readable \texttt{exclusion\_reason}, of which five forms occur: \texttt{number\_only\_in\_figure}, \texttt{figure\_only}, \texttt{figure\_only\_logistic\_prior}, \texttt{no\_numeric\_estimate}, and \texttt{different\_unit\_of\_analysis}; these account for the 5 of 99 papers outside the panel. Values are never interpolated or adjusted toward the existing anchor, and a paper's different-but-equivalent unit is converted explicitly and noted. Where a panel is empty---possible when an MCM answer is prose or figure only---the task falls back to the single reproduced anchor, and a missing artifact scores $0$, so the denominator is always the fixed 18 tasks.

A panel also makes disagreement between humans visible instead of hiding it behind one anchor. On \texttt{mcm-2024-c-tennis-momentum}, 10 of 11 outstanding papers conclude that momentum is supported while the reproduced anchor---itself an O-award report---concludes it is not; on that task a model scores $0.909$ if it answers with the majority and $0.091$ if it matches the anchor, so the score reflects which valid statistical test the model chose rather than modeling quality. We mark such \emph{literature-split} tasks explicitly and exclude them from any headline ranking, while keeping them in the panel: they are the clearest evidence that a single-anchor score would have been arbitrary.

Relative to the single-anchor board, this changes \roeval{} into a \emph{multi-reference} gain index against the DeepSeek~V4.1~Flash baseline, anchored at $1.0$:
\begin{equation}
\boeval \;=\; 1 + \mathrm{clip}\!\left(\frac{O_t - f_t}{1 - f_t},\, -1,\, +1\right),
\end{equation}
with $f_t$ the Flash score on task $t$ and $O_t$ the model's, averaged over tasks; when the remaining headroom $1-f_t$ is below $0.10$ we fall back to $1 + \mathrm{clip}(O_t - f_t, \pm 0.10)$, and a missing artifact scores $-1$. A value of $1.0$ therefore means ``level with Flash'', $1.08$ means ``closed 8\% of the remaining gap to the human panel''; it is a ratio, not a percentage, and is not comparable to \oeval{}.

\subsection{Feasibility verification: two layers}
\label{sec:verification}
Proximity to an anchor does not imply that an answer is \emph{physically admissible}: a submission can match a reference value on one field while violating a stated constraint on another. \bench{} verifies feasibility independently of scoring, in two layers with different trust bases.

\paragraph{Layer 1: a programmatic gate over six decidable forms.}
Every task registers one feasibility check, and the registry covers \textbf{18/18} tasks; every hard-error branch falls into one of six decidable forms (Appendix Table~\ref{tab:invariants}). Checks are either structural/domain bounds (rates, correlations, probabilities, non-negative counts, distributions summing to 100, peak $\ge$ average) or constraint replay (heliostat rated power $\ge 60$~MW and installation height $\in[2,6]$~m, dragon-dance head speed $\times$ amplification ratio $\le 2$~m/s, a demand-capped revenue ceiling, a union duration $\le$ the sum of its parts). Three contracts keep the gate honest: \textbf{coverage} (a task cannot ship ungated), \textbf{oracle self-consistency} (the O-award reference must pass its own check), and \textbf{discrimination} (synthetic and real-field violations must be rejected, while an empty artifact warns rather than hard-fails). Four constraints that need an independent replay rather than a field check are recorded as whole-task penalties (Appendix~\ref{app:scoring}).

\paragraph{Layer 2: Lean certificates.}
To make the gate's own correctness checkable, Layer~2 re-proves the same invariants as Lean~4 theorems, so the audit rests on the kernel. It generates \textbf{624 obligations across all 18 tasks}, discharged by \texttt{decide} or \texttt{omega}. \textbf{No Mathlib is used}---the propositions are decidable on exact integers---and values are scaled to integers, so there is no floating-point tolerance; the one exception, a composite index, is \emph{declared} in the predicate because the tolerance is itself part of the audited assumption. A trust audit (\texttt{\#print axioms}) reports only \texttt{[propext, Quot.sound]} or none, with no \texttt{sorryAx} and no \texttt{native\_decide} (grep-checked). A discrimination suite mutates one numeric leaf at a time and requires the build to fail \emph{and} to cite the field; \textbf{ten mutations across all six forms} are rejected. Two soft warnings become proofs: a tourism acceptance index is \emph{refuted} for the unit convention, and a sustainability score fits \emph{neither} convention within a declared budget.

\paragraph{Verification scope.}
The two layers decide every feasibility question that a structured artifact can settle without a model of the physics (Appendix Table~\ref{tab:abc}): unit/value ranges (A) and internal consistency (B). Whether the reported value reflects real physical truth (C)---does the field really produce $60$~MW?---is a property of the world rather than of the artifact, so \bench{} verifies it where the problem admits an independent replay. Within this scope the layers are exact: any obligation whose evidence is absent is \emph{reported} as a witnessed gap rather than silently satisfied, so the audit never overstates what it has proved.

%% file: contents/4.Experiment.tex
\section{Experiments}
\label{sec:results}

\subsection{Setup}
\label{sec:setup}
We evaluate \textbf{seven multimodal model/harness entries}, whose harness can open the cropped figures. Each entry is run with \textbf{three seeds per task} and reported as a mean with standard deviation; where a model was run at two reasoning tiers we keep the higher-scoring variant. Every row is scored with the same scorer and current anchors, so the board shares one scoring definition (Appendix~\ref{app:settings}).

The prompt is the task's instruction file, passed verbatim: the contest statement, the cited figures inline, and a fixed ``Required Benchmark Artifact'' block naming the output path (Appendix~\ref{app:prompts}). Tasks run in Docker containers---\texttt{python:3.12-slim} with a fixed scientific stack, \textbf{internet disabled}, model traffic through a local retry proxy---so an agent cannot fetch solutions or install packages at run time; limits are 2 CPUs, 8~GB, no GPU, and a 4-hour timeout (Appendix~\ref{app:environment}). Each multimodal entry is run three times per task and reported as a mean with standard deviation; the one well-posed task, dragon-dance, is additionally probed four times, where the answers agreed to machine precision (\texttt{std}=$4.6\times10^{-16}$).

\subsection{Main result: the multimodal board}
\label{sec:main}
Table~\ref{tab:main} is the main result: the seven multimodal entries on all 18 tasks, each a three-seed mean ($\pm$ std); the per-entry configuration is in Appendix Table~\ref{tab:full}. \oeval{} is the primary metric: the \emph{multi-reference} score of Section~\ref{sec:multi-ref}, i.e.\ the mean over the human paper panel of how close the model is to each accepted answer. \boeval{} is the secondary, relative index of that same score against the DeepSeek~V4.1~Flash baseline, fixed at $1.000$ cell-by-cell: $1.08$ means the model closed $8\%$ of Flash's remaining gap to the panel, so it is read against the baseline and is not a percentage. Five findings stand out.

\begin{enumerate}[leftmargin=1.5em,itemsep=2pt]
\item \textbf{The newest flagships lead.} GPT-6 Astra + codex ($64.58 \pm 2.93$) and Fable 5.1 + claude-code ($63.11 \pm 1.33$) top the board, about 4.8 and 3.4 points above GPT-5.5 xhigh + codex ($59.75 \pm 1.55$), and both clear the Flash baseline by $13.2\%$ and $8.1\%$ of the remaining gap respectively.
\item \textbf{Cost buys a real but bounded gap.} The two leaders cost \$246--\$273 per run, while the cheapest strong entry, DeepSeek-Flash + opencode, reaches $59.81 \pm 1.25$ at \$2.54---about 4.8 points behind the leader at roughly $1/100$ the cost.
\item \textbf{Cheap open models cluster below the flagships.} Among the low-cost entries DeepSeek-Flash (59.81), Kimi K3 (58.23), GLM-5.3 (57.78), and Qwen3.8 Flash (52.66) span about 7 points, and only the two flagships exceed the baseline index ($1.132$ and $1.081$); DeepSeek-Flash defines it at $1.000$ and every other entry falls below.
\item \textbf{A small gap is not yet a resolved ranking.} Per-seed deviations reach $2.93$~pp, so the $\sim1.5$~pp separation between the two leaders and between GPT-5.5 and Flash sits inside one standard deviation; we report the order but do not claim a significant difference between adjacent rows.
\item \textbf{More inference compute is not monotone.} Running DeepSeek-Flash at its \emph{maximum} reasoning tier drops its multi-reference score to $48.11\%$ and its \boeval{} to $0.864$, well below the default tier's $59.81$ and $1.000$, at higher cost: a larger deliberative budget does not simply buy accuracy on these tasks.
\end{enumerate}

\begin{table*}[t]
\centering
\small
\setlength{\tabcolsep}{6pt}
\caption{Main results: the seven multimodal entries, 18 tasks, sorted by \oeval{}. ``Art.'': tasks with a scoreable artifact; cost: USD per 18-task pass. \oeval{} is the multi-reference panel score (\%) and \boeval{} the same score relative to the DeepSeek~V4.1~Flash baseline ($1.000$ = level with Flash), both three-seed means ($\pm$ std).}
\label{tab:main}
\begin{tabular}{@{}lcccc@{}}
\toprule
Model + harness & Art. & \oeval{}(\%) & \boeval{} & Cost (USD) \\
\midrule
GPT-6 Astra + codex          & 18 & $64.58 \pm 2.93$ & $1.132 \pm 0.02$ & 246.41 \\
Fable 5.1 + claude-code      & 18 & $63.11 \pm 1.33$ & $1.081 \pm 0.05$ & 272.80 \\
DeepSeek-Flash + opencode    & 18 & $59.81 \pm 1.25$ & $1.000$          & 2.54 \\
GPT-5.5 xhigh + codex        & 18 & $59.75 \pm 1.55$ & $0.998 \pm 0.01$ & 99.37 \\
Kimi K3 + opencode           & 18 & $58.23 \pm 1.12$ & $0.953 \pm 0.04$ & 23.26 \\
GLM-5.3 + opencode           & 18 & $57.78 \pm 1.30$ & $0.941 \pm 0.05$ & 28.15 \\
Qwen3.8 Flash + opencode     & 18 & $52.66 \pm 1.17$ & $0.803 \pm 0.04$ & 8.30 \\
\bottomrule
\end{tabular}
\end{table*}

\begin{figure*}[t]
\centering
\includegraphics[width=\textwidth]{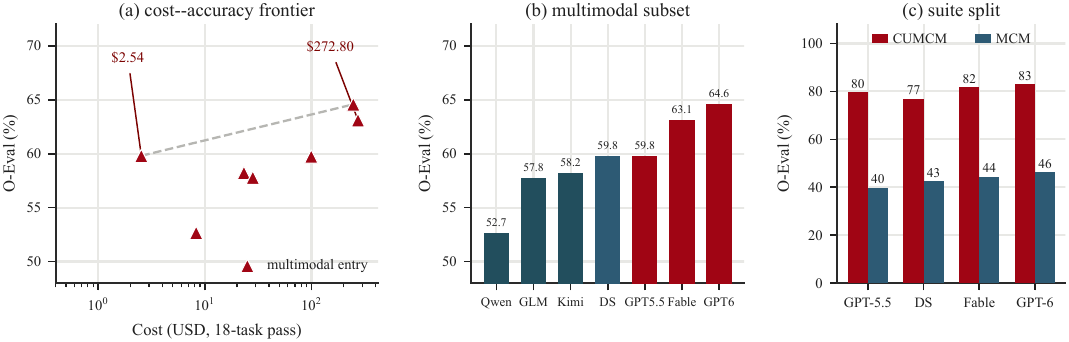}
\caption{Left: multi-reference \oeval{} versus cost (log scale)---the Pareto frontier is compressed, with near-frontier scores at two orders of magnitude lower cost. Centre: \oeval{} per entry. Right: \oeval{} by suite, showing the CUMCM--MCM gap.}
\label{fig:results}
\end{figure*}

\paragraph{Hard gate and cost.}
The gate removes exactly the cells whose answers violate a stated constraint and leaves the ranking otherwise stable, so it is surgical rather than a blanket penalty. Cost is dominated by per-token price, not usage: GPT-5.5's \$99.37 is mostly cache-read input and output at roughly $33\times$/$50\times$ DeepSeek~V4.1~Flash's rates, despite comparable token counts.

\paragraph{Harness sensitivity.}
Each model is driven by an agent harness, so a natural question is whether the score reflects the model or the harness. Table~\ref{tab:harness} holds the model fixed and varies the harness, using DeepSeek-Flash and four harnesses. The score is essentially invariant: \oeval{} moves only $0.07$~pp across all four ($59.81$--$59.88$), far inside the per-seed deviation, and the relative index moves in the same narrow band. A model's own harness fits it marginally better---DeepSeek's native harness gives the highest \oeval{} ($59.88$) and \boeval{} ($1.092$)---but the difference is small. The board therefore measures the model far more than the harness it is wrapped in.

\begin{table}[t]
\centering
\small
\setlength{\tabcolsep}{5pt}
\caption{Harness sensitivity, holding the model fixed (DeepSeek-Flash, 18 tasks, three seeds). Multi-BO is $1.000$ for the opencode reference harness by definition.}
\label{tab:harness}
\begin{tabular}{@{}lcc@{}}
\toprule
Harness & \oeval{}(\%) & \boeval{} \\
\midrule
opencode            & $59.81 \pm 1.25$ & $1.000$ \\
claude-code         & $59.83 \pm 1.13$ & $1.081 \pm 0.05$ \\
codex               & $59.84 \pm 1.21$ & $1.085 \pm 0.03$ \\
DeepSeek (native)   & $59.88 \pm 1.20$ & $1.092 \pm 0.03$ \\
\bottomrule
\end{tabular}
\end{table}

\subsection{Analysis}
\label{sec:analysis}
Per-task \oeval{} for four representative entries is in Appendix Table~\ref{tab:pertask}. Three patterns stand out. \textbf{(a) The CUMCM--MCM gap is large and model-dependent.} For the weaker entries, CUMCM cells sit near 100\% while MCM cells sit between 0\% and 60\%: the engineering problems have well-defined final numbers most models reach, whereas MCM problems are more open and their anchors more contestable. The strongest entries narrow the gap only slightly (GPT-6 82.9 on CUMCM vs.\ 46.2 on MCM; GPT-5.5 79.7 vs.\ 39.8). A sizeable share of tasks is now solved near-perfectly---expected of a current benchmark---and the board's discrimination comes from the open-ended MCM problems. \textbf{(b) Two tasks are all-or-nothing.} On \texttt{mcm-2024-b-submersible} every entry scores 100\%, while on \texttt{mcm-2024-c-tennis} even the flagships score below 40\%. Such tasks either fall inside or outside the panel's tolerance entirely, so they separate regimes of models but not models within one. \textbf{(c) Many low scores reflect reference scale/scope, not modeling error.} Several tasks contain a metric whose model/panel ratio exceeds $5\times$: a host-fish index reported near 1080 by some papers while models give a normalised $[0,1]$ quantity; a sustainability score near 1.007 against models reporting a 0--100 utility; a $6\times6$ grid against models discretising the same region into hundreds of cells; and a $72.67\!\to\!73.52$ band in which the paper's own baseline scores zero.

\paragraph{Scoring pitfalls we found and fixed.}
An anchor-based benchmark is only as credible as its anchors, so we audited every scoring rule and anchor and report what the audit changed (Appendix~\ref{app:scoring}). Two gate false positives were removed without moving any score, and degenerate-band widening affected one cell ($+0.35$\,pp). Excluding two non-comparable-unit metrics raised \emph{every} entry by 0.05--2.62\,pp---the signature of a systematic scoring effect rather than a model difference. Most consequentially, the audit flagged the dragon-dance anchor as physically infeasible and we corrected it (Section~\ref{sec:main}).

\paragraph{What the feasibility layers caught.}
The gate hard-invalidated 7 cells, concentrated on heliostat (rated power below 60~MW, or a mirror below the 2~m minimum) plus single cells for smoke-screen, Juneau, and Olympic. The Lean layer turns two soft warnings into proofs and proves non-vacuity by accepting a corrected artifact under the predicate that rejects the reported one. On the one numerically well-posed task the runs agree to machine precision, and across the board the per-seed deviations stay small relative to the spread between entries (Table~\ref{tab:main}), so rankings do not depend on the seed.

\subsection{From Findings to Model Development}
\label{sec:develop}
\bench{} is meant to improve models, not only rank them. Three levers follow from the findings above, and each is trainable with signals \bench{} already produces.

\textbf{Verified reward for RL.} The feasible, oracle-normalised score is an executable, reproducible reward---every submission passes the gate and the Lean layer. Rejection-sampling (keep and rank gate-accepted trajectories by \oeval{}) works today; group-relative policy optimisation with the gate as a hard multiplier teaches the model to satisfy the physics before chasing the reference. The dragon-dance case motivates this: twelve models computed the right peak but were penalised, so what they lack is self-checking---exactly what a verifier in the reward provides.

\textbf{Targets against the observed failures.} Finding (a)---the open-ended MCM suite separates quality while CUMCM saturates---says to train assumptions, formulation, and sensitivity rather than closed-form solving; finding (c)---many losses are unit and scale mismatches---says to state and verify units, ranges, and schema before emitting a number. Both are rewarded directly by the artifact contract.

\textbf{Feasibility the artifact can check.} The dragon-dance audit exposes the failure a value-range check cannot: a consistent answer that is physically inadmissible because the artifact omits the evidence (no per-mirror layout, no peak-neighbourhood sample). Training models to emit that \emph{witness} raises the score and makes them checkable, reaching level C.

\paragraph{Case studies.}
Three roles---a top proprietary agent, the oracle, and the low-cost open agent---are compared on three problems in Appendix~\ref{app:cases}. On \emph{dragon-dance} all three converge while the paper's value is the infeasible outlier, so the \emph{models} were right and our anchor was wrong. On \emph{crop-planting} the paper's constraints are jointly infeasible, so we anchor on a reproducible exact MILP optimum. On \emph{Juneau} the sustainability score is reported on different unit scales, which is why the task \emph{schema}, not the scorer, must pin a field's range.

%% file: contents/5.Conclusion.tex
\section{Conclusion}
\label{sec:conclusion}

We presented \bench{}, the first agentic and verifiable terminal benchmark of 18 real mathematical-modeling problems, distinguished by \emph{direction-aware, one-sided scoring} that never penalises beating the anchor, and \emph{two-layer feasibility verification} combining a gate over six decidable invariant forms with 624 kernel-checked Lean obligations.

Scoring each task against a panel of human outstanding papers, the newest flagships lead across seven multimodal entries (GPT-6 Astra at \oeval{} $64.58\%$, Fable 5.1 at $63.11\%$), while the \$2.54 open DeepSeek-Flash holds $59.81\%$ at $1/100$ the cost. We documented the scoring pitfalls we fixed, and showed a case where the benchmark itself was wrong and the models were right.

The two-layer design is the transferable idea: a fast gate covers every task and a proof assistant makes the audit's own correctness machine-checkable, while stating what neither can answer. Because \bench{} gives a reproducible reward, its trajectories are also training material, and Section~\ref{sec:develop} turns the findings into directions.

%% file: contents/Acknowledgements.tex
\section*{Acknowledgements}
We thank OpenRefinery (\url{https://openrefinery.ai/}) for providing the LLM API access and evaluation support that enabled our experiments with GPT-6-Astra and Claude-Fable-5.1. In particular, we thank Kai Du and Ding Fan of OpenRefinery for their assistance. Their infrastructure and support were instrumental in conducting the model evaluations reported in this paper. Any opinions, findings, and conclusions expressed here are those of the authors and do not necessarily reflect the views of OpenRefinery.

%% file: contents/Limitations.tex
\section*{Limitations}
\label{sec:limitations}
\bench{} normalises each task against a reproducible Outstanding-award solution, with per-anchor provenance so any reference can be re-derived; because modeling admits no single correct answer, the natural extension is a multi-paper panel reporting a spread over accepted solutions. The two verification layers decide every feasibility question a structured artifact can settle (unit/value ranges and internal consistency); external physical truth is a property of the world, so it is checked where a problem admits an independent replay, and any obligation whose evidence is missing is reported rather than silently passed. Panel size varies with the number of outstanding reports per problem (1 to 18 papers), and where humans themselves disagree---as on the tennis momentum test, where 10 of 11 papers reach the opposite conclusion from the anchor---we mark the task literature-split and keep it out of headline rankings. Scores are three-seed means, and the task shell carries over to engineering and cross-disciplinary problems.

\section*{Ethical Considerations}
\label{sec:ethics}
Statements, attachments, and figures are official CUMCM and MCM material used for research evaluation; we do not redistribute Outstanding-award solutions, and reproduction code derives from a public corpus with per-task provenance. Users should confirm the licence terms for their own redistribution. \bench{} measures and improves the \emph{modeling workflow} of LLM agents rather than automating work presented as a team's own, and entirely automated evaluation keeps the cost of broad participation low.

%% file: contents/LLM_usage.tex
\section*{Use of Large Language Models}
\label{sec:llm-usage}
LLMs served as the \emph{systems under test}---every entry in Table~\ref{tab:main} is an LLM agent, which is the object of measurement---and as \emph{writing assistants} that helped draft prose and check the internal consistency of \LaTeX{} and numbers. Benchmark construction, scoring rules, feasibility checks, and Lean obligations were specified, implemented, and verified by the authors, and all numerical results are produced by the released scripts.

%% file: contents/Appendix/0.Appendix_overview.tex
\section*{Appendix Overview}

This supplementary material makes the main paper's construction and claims auditable. Appendix~\ref{app:dataset} gives the dataset composition, the per-task table, the delivered directory structure, and an explicit statement that \bench{} has no train/validation/test split. Appendix~\ref{app:environment} records the software and hardware environment and the network isolation policy. Appendix~\ref{app:settings} lists the per-entry configuration, the agent command templates, and the retry/artifact-recovery details. Appendix~\ref{app:baselines} disambiguates the three objects called ``baseline''. Appendix~\ref{app:prompts} reproduces the prompt template and the harness-wrapping layer. Appendix~\ref{app:cases} gives the full numerics behind the three case studies. Appendix~\ref{app:scoring} gives the scoring formulas, the excluded metrics, and the hard-gate decisions. Appendix~\ref{app:lean} catalogues the Lean obligations, the trust audit, and the discrimination suite.

\begin{itemize}[leftmargin=1.4em,itemsep=1pt]
    \item Appendix~\ref{app:dataset}: dataset composition, per-task table, delivered structure, and split
    \item Appendix~\ref{app:environment}: host, agent container, verifier container, resource limits, network isolation
    \item Appendix~\ref{app:settings}: per-entry configuration, command templates, retries, cost accounting
    \item Appendix~\ref{app:baselines}: flash baseline, Outstanding anchor, legacy endpoints, anchor caveats
    \item Appendix~\ref{app:prompts}: prompt structure, artifact block, harness wrapping, version caveat
    \item Appendix~\ref{app:cases}: full numeric case-study tables
    \item Appendix~\ref{app:scoring}: metric panel, formulas, excluded metrics, hard gates
    \item Appendix~\ref{app:lean}: obligation counts, representative obligations, trust audit, discrimination suite
\end{itemize}

\input{contents/Appendix/1.Dataset}
\input{contents/Appendix/2.Environment}
\input{contents/Appendix/3.Settings}
\input{contents/Appendix/4.Baselines}
\input{contents/Appendix/5.Prompts}
\input{contents/Appendix/6.Case_studies}
\input{contents/Appendix/7.Scoring_details}
\input{contents/Appendix/8.Lean_catalog}

%% file: contents/Appendix/1.Dataset.tex
\section{Dataset Details}
\label{app:dataset}

\subsection{Composition}

\begin{table*}[t]
\centering
\small
\setlength{\tabcolsep}{12pt}
\caption{The 18 \bench{} tasks and their O-award anchors. ``Final Q'' is the contest question whose numeric core is scored; oracle coverage is 18/18.}
\label{tab:tasks}
\begin{tabular}{@{}llll@{}}
\toprule
Slug & Contest & Topic & Final Q \\
\midrule
\texttt{cumcm-2023-a-heliostat}  & CUMCM 2023 A & Heliostat field design & Q3 \\
\texttt{cumcm-2023-b-multibeam}  & CUMCM 2023 B & Multibeam survey lines & Q4 \\
\texttt{cumcm-2023-c-vegetable}  & CUMCM 2023 C & Vegetable pricing and replenishment & Q4 \\
\texttt{cumcm-2024-a-dragon}     & CUMCM 2024 A & Bench-dragon kinematics & Q5 \\
\texttt{cumcm-2024-b-production} & CUMCM 2024 B & Production decision & Q4 \\
\texttt{cumcm-2024-c-crop}       & CUMCM 2024 C & Crop-planting strategy & Q1 \\
\texttt{cumcm-2025-a-smoke}      & CUMCM 2025 A & UAV smoke-screen strategy & Q5 \\
\texttt{cumcm-2025-b-sic}        & CUMCM 2025 B & SiC epi-layer thickness & Q3 \\
\texttt{cumcm-2025-c-nipt}       & CUMCM 2025 C & NIPT timing / abnormality & Q4 \\
\texttt{mcm-2023-a-plant}        & MCM 2023 A & Drought plant communities & final \\
\texttt{mcm-2023-b-maasai}       & MCM 2023 B & Maasai Mara land use & final \\
\texttt{mcm-2023-c-wordle}       & MCM 2023 C & Predicting Wordle results & final \\
\texttt{mcm-2024-a-lamprey}      & MCM 2024 A & Sex ratios / resource supply & final \\
\texttt{mcm-2024-b-submersible}  & MCM 2024 B & Submersible search & Q4 \\
\texttt{mcm-2024-c-tennis}       & MCM 2024 C & Momentum in tennis & Q5 \\
\texttt{mcm-2025-a-stair}        & MCM 2025 A & Stair wear / footfall & Q8 \\
\texttt{mcm-2025-b-juneau}       & MCM 2025 B & Sustainable tourism (Juneau) & Q3 \\
\texttt{mcm-2025-c-olympic}      & MCM 2025 C & Olympic medal models & final \\
\bottomrule
\end{tabular}
\end{table*}

\bench{} has \textbf{18 tasks}, listed in Table~\ref{tab:tasks} (main text). The composition is CUMCM (Chinese) 2023, 2024, 2025 $\times$ problems A, B, C $=$ 9 tasks, plus MCM (English) 2023, 2024, 2025 $\times$ problems A, B, C $=$ 9 tasks. Each task is a full contest problem with several sub-questions; only the numeric core of the \emph{final} question is scored. Problem~A is typically physical/engineering, B optimization/data, and C data-science, so taking all three of each year avoids topical cherry-picking.

\subsection{Per-task summary}
Table~\ref{tab:dataset} summarises each task. ``\#Fig.'' counts unique figures cropped from the official PDF; ``\#Metric'' is the number of scored metrics.

\begin{table*}[h]
\centering
\setlength{\tabcolsep}{5pt}
\small
\setlength{\tabcolsep}{5pt}
\caption{Per-task summary. ``Final Q'' is the sanctioned final question.}
\label{tab:dataset}
\begin{tabular}{@{}lllcc@{}}
\toprule
Slug & Final Q & Source & \#Fig. & \#Metric \\
\midrule
cumcm-2023-a-heliostat   & Q3 & CUMCM 2023 A & 2 & 5 \\
cumcm-2023-b-multibeam   & Q4 & CUMCM 2023 B & 8 & 4 \\
cumcm-2023-c-vegetable   & Q4 & CUMCM 2023 C & 0 & 2 \\
cumcm-2024-a-dragon      & Q5 & CUMCM 2024 A & 5 & 2 \\
cumcm-2024-b-production  & Q4 & CUMCM 2024 B & 1 & 6 \\
cumcm-2024-c-crop        & Q1 & CUMCM 2024 C & 0 & 2 \\
cumcm-2025-a-smoke       & Q5 & CUMCM 2025 A & 0 & 2 \\
cumcm-2025-b-sic         & Q3 & CUMCM 2025 B & 2 & 2 \\
cumcm-2025-c-nipt        & Q4 & CUMCM 2025 C & 0 & 2 \\
mcm-2023-a-plant         & final & MCM 2023 A & 0 & 3 \\
mcm-2023-b-maasai        & final & MCM 2023 B & 0 & 5 \\
mcm-2023-c-wordle        & final & MCM 2023 C & 2 & 2 \\
mcm-2024-a-lamprey       & final & MCM 2024 A & 0 & 3 \\
mcm-2024-b-submersible   & Q4 & MCM 2024 B & 0 & 2 \\
mcm-2024-c-tennis        & Q5 & MCM 2024 C & 0 & 4 \\
mcm-2025-a-stair         & Q8 & MCM 2025 A & 1 & 3 \\
mcm-2025-b-juneau        & Q3 & MCM 2025 B & 0 & 6 \\
mcm-2025-c-olympic       & final & MCM 2025 C & 0 & 5 \\
\midrule
\textbf{Total} & & & \textbf{21} & \textbf{60}$^{\dagger}$ \\
\bottomrule
\end{tabular}
\end{table*}

$^{\dagger}$60 scored metrics across the 18 tasks; one further Wordle metric is declared at weight~0 and excluded from scoring. The counts are verified against the released tasks; the 21 figures are unique after de-duplication and are delivered twice---once inline in the instruction file and once under the container figure path. Data sources are the official contest material: CUMCM statements under \texttt{cumcm/problems/\{year\}/\{X\}.md} and MCM under \texttt{mcm/problems/\{year\}/\{X\}.md}; Outstanding-paper PDFs/OCR used for anchors come from the same corpus's \texttt{outstanding/} tree, and every task records its paper identifier, the PDF/OCR source paths, and its \texttt{reproduction\_scope}.

\subsection{Per-task multi-reference scores}
\begin{table*}[h]
\centering
\small
\setlength{\tabcolsep}{8pt}
\caption{Per-task multi-reference \oeval{} (\%) for four representative entries. Each column's mean equals that entry's main-table \oeval{}, so the per-task values reconcile with the board.}
\label{tab:pertask}
\begin{tabular}{@{}lcccc@{}}
\toprule
Task & GPT-6 Astra & Fable 5.1 & GPT-5.5 xhigh & DeepSeek-Flash \\
\midrule
\texttt{cumcm-2023-a-heliostat} & 86.2 & 80.7 & 80.5 & 80.2 \\
\texttt{cumcm-2023-b-multibeam} & 94.0 & 80.5 & 91.7 & 59.3 \\
\texttt{cumcm-2023-c-vegetable} & 64.3 & 64.3 & 64.3 & 64.3 \\
\texttt{cumcm-2024-a-dragon} & 83.4 & 83.4 & 83.4 & 83.4 \\
\texttt{cumcm-2024-b-production} & 95.7 & 80.5 & 87.2 & 87.2 \\
\texttt{cumcm-2024-c-crop} & 88.2 & 83.5 & 99.7 & 63.8 \\
\texttt{cumcm-2025-a-smoke} & 100.0 & 100.0 & 100.0 & 100.0 \\
\texttt{cumcm-2025-b-sic} & 88.2 & 86.7 & 80.3 & 78.7 \\
\texttt{cumcm-2025-c} & 46.4 & 77.0 & 29.7 & 76.4 \\
\texttt{mcm-2023-a-plant} & 21.3 & 24.6 & 27.7 & 27.7 \\
\texttt{mcm-2023-b-maasai} & 34.5 & 35.1 & 43.1 & 43.1 \\
\texttt{mcm-2023-c} & 46.1 & 49.6 & 40.9 & 56.4 \\
\texttt{mcm-2024-a} & 47.1 & 43.8 & 10.7 & 60.9 \\
\texttt{mcm-2024-b-submersible} & 100.0 & 100.0 & 100.0 & 100.0 \\
\texttt{mcm-2024-c-tennis} & 36.4 & 36.4 & 9.1 & 9.1 \\
\texttt{mcm-2025-a-stair} & 21.4 & 13.1 & 12.9 & 12.8 \\
\texttt{mcm-2025-b-juneau} & 42.9 & 42.9 & 42.9 & 42.9 \\
\texttt{mcm-2025-c-olympic} & 66.4 & 54.1 & 71.2 & 30.1 \\
\midrule
\textbf{Average} & 64.58 & 63.11 & 59.75 & 59.81 \\
\bottomrule
\end{tabular}
\end{table*}

\subsection{Delivered structure}
Each task directory contains an instruction file (the prompt), a metadata file (limits), an environment directory (agent image plus data), a tests directory (verifier, scoring config, anchor), and a solution directory (oracle). Inside the environment data directory are the statement copy, the output schema (field names and types only), a source manifest mapping each visible file to its corpus path, the verbatim attachments under flat filenames with an index, a full-path mirror of the original corpus, and the cropped figures for the six tasks that have them.

\subsection{Split}

\begin{figure}[t]
\centering
\includegraphics[width=0.66\linewidth]{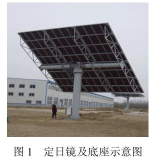}
\caption{An official statement figure, cropped from the CUMCM 2023~A PDF and delivered inline in the prompt. 21 such figures across 6 tasks are load-bearing: the final answer depends on reading them.}
\label{fig:official-fig}
\end{figure}

\textbf{There is no train/validation/test split.} \bench{} is a pure evaluation set: the 18 tasks are all evaluation tasks, evaluated uniformly, with no held-out partition and no per-task difficulty tier. (One task's oracle model contains an in-task train/test split for the Wordle problem; that is part of that problem's modeling content and is unrelated to benchmark partitioning.)

%% file: contents/Appendix/2.Environment.tex
\section{Software and Hardware Environment}
\label{app:environment}

\subsection{Host and framework}
The evaluation harness runs from a macOS host (arm64). It orchestrates per-task Docker containers; the enforced resource limits are container limits, not host limits.

\subsection{Agent container}
All 18 tasks use the same base and dependency set. The base image is \texttt{python:3.12-slim} plus \texttt{uv}. The Python environment at \texttt{/opt/venv} pins NumPy 2.0.2, pandas 2.2.2, SciPy 1.14.1, scikit-learn 1.5.2, Matplotlib 3.9.2 (\texttt{MPLBACKEND=Agg}), openpyxl 3.1.5, and xlrd 2.0.1. Writable paths are \texttt{/root/data} (read-only inputs), \texttt{/root/results} (the scored artifact), and \texttt{/logs/agent}.

\subsection{Verifier container}
The verifier runs in a \emph{separate} container built from a minimal \texttt{python:3.12-slim} image installing only \texttt{pytest}. It has no network and no access to the agent's working directory beyond the copied artifacts. The verifier timeout is 180~s.

\subsection{Resource limits}
The limits are identical for all tasks: 2 CPUs, 8192~MB memory, 20480~MB storage, 0 GPUs. The build timeout is 900~s and the agent timeout 14400~s (4~h), also stated at the end of each prompt. The recorded runs used one concurrent trial and no retries.

\subsection{Network isolation}
Every task disables internet access. Model traffic is routed through a local retry proxy, so an agent can call the model API but cannot fetch solutions, download packages, or install tooling at run time. Because installation is impossible after a trial starts, CLI agents (\texttt{codex}, \texttt{opencode}, \texttt{claude-code}) must be present in the image; the third-party operator's runs pre-installed the CLIs at the end of the Dockerfile and allow-listed only the model gateway, leaving the statement, data, and verifier unchanged.

\subsection{Frozen versions}
For reproducibility we pin the harness version, the Python dependency set above, and the local pricing snapshot used for cost, so that the reported costs are stable across reruns.

%% file: contents/Appendix/3.Settings.tex
\section{Experimental Settings}
\label{app:settings}

\subsection{Per-entry configuration and results}
Table~\ref{tab:full} lists the seven multimodal entries with the harness and reasoning tier actually passed and their full scores. Multimodal status means the harness exposes an image channel. Every entry is run with three seeds per task.

\begin{table*}[h]
\centering
\small
\setlength{\tabcolsep}{8pt}
\caption{Multimodal board, 18 tasks, sorted by \oeval{}. ``Art.'' is the number of tasks with a scoreable artifact; \oeval{} is the multi-reference panel score and \boeval{} the same score relative to the DeepSeek~V4.1~Flash baseline, both three-seed means ($\pm$ std).}
\label{tab:full}
\begin{tabular}{@{}llccc@{}}
\toprule
Model & Harness / tier & Art. & \oeval{}(\%) & \boeval{} \\
\midrule
GPT-6 Astra      & codex CLI / max       & 18 & $64.58 \pm 2.93$ & $1.132 \pm 0.02$ \\
Fable 5.1        & claude-code CLI / max & 18 & $63.11 \pm 1.33$ & $1.081 \pm 0.05$ \\
GPT-5.5 xhigh    & codex CLI / xhigh     & 18 & $59.75 \pm 1.55$ & $0.998 \pm 0.01$ \\
DeepSeek-Flash   & opencode CLI / high   & 18 & $59.81 \pm 1.25$ & $1.000$ \\
Kimi K3          & opencode CLI / high   & 18 & $58.23 \pm 1.12$ & $0.953 \pm 0.04$ \\
GLM-5.3          & opencode CLI / high   & 18 & $57.78 \pm 1.30$ & $0.941 \pm 0.05$ \\
Qwen3.8 Flash    & opencode CLI / high   & 18 & $52.66 \pm 1.17$ & $0.803 \pm 0.04$ \\
\bottomrule
\end{tabular}
\end{table*}

\subsection{Agent command templates}
The task statement is passed to the agent as its prompt. Representative commands from the recorded trials:
\begin{lstlisting}[basicstyle=\ttfamily\scriptsize]
# codex
codex exec --dangerously-bypass-approvals-and-sandbox --skip-git-repo-check \
  --model gpt-5.5 --json --enable unified_exec \
  -c model_reasoning_effort=xhigh -- '<instruction.md body>'

# opencode
opencode --model=openai/deepseek-flash run --format=json --thinking \
  --dangerously-skip-permissions -- '<instruction.md body>'
\end{lstlisting}
Each CLI harness receives the statement as its positional prompt; the harness prepends its own generic system prompt before it.

\subsection{Retries, concurrency, and artifact recovery}

All entries used one concurrent trial and no automatic retries; each of the three seeds is an independent pass over the 18 tasks. Two subtleties are disclosed. First, \textbf{misfiled artifacts}: two tasks wrote the result with a hyphen instead of the required underscore, so the harness did not collect it; we replayed the recorded run's scripts to reproduce the artifact and re-scored it. Second, \textbf{third-party runs}: the GPT-6 Astra and Fable~5.1 entries were run by independent operators; because their verifier used a different reward formula, we re-scored their saved artifacts with \emph{our} scorer and current anchors so that all entries share one scoring definition. Where an artifact was not packaged, we reconstructed the scored fields from the saved per-metric records, verified to reproduce the operator's own numbers exactly on the tasks where the full artifact was present.

\subsection{Cost accounting}
Cost is computed from measured token usage against a \emph{fixed} price snapshot, with two conservative adjustments stated in the report: reasoning tokens that a provider bills outside its completion counter are added to output, and where a provider charges a peak/off-peak rate, the peak rate is used. For the two third-party entries, cost is their own measured per-run USD. Because the pricing file is local and fixed, costs do not drift between reruns.

%% file: contents/Appendix/4.Baselines.tex
\section{Baseline Information}
\label{app:baselines}

The word ``baseline'' is overloaded in this benchmark; we distinguish three distinct objects so that the numbers are unambiguous.

\subsection{The flash baseline (score normalisation)}
\roeval{} is computed against a \emph{single reference submission}, the DeepSeek~V4~Flash baseline run on the same tasks. It defines the zero point: \roeval{} normalises the model--baseline gain by the oracle--baseline gap. It is a model submission, not a curated baseline panel.

\subsection{The Outstanding anchor (score ceiling)}
Each task's ceiling is the reproduction of one Outstanding-award paper, which is the value that scores \oeval{}$=100\%$. Table~\ref{tab:app-anchors} lists the anchor paper identifier for each task. Each anchor is a runnable script that reads the official attachments and emits the artifact JSON, plus a committed expected result recording its values; coverage is 18/18.

\begin{table*}[h]
\centering
\setlength{\tabcolsep}{5pt}
\small
\setlength{\tabcolsep}{6pt}
\caption{Outstanding-award anchor per task.}
\label{tab:app-anchors}
\begin{tabular}{@{}llll@{}}
\toprule
Slug & Paper id & Slug & Paper id \\
\midrule
cumcm-2023-a-heliostat   & A175    & mcm-2023-a-plant         & 2309229 \\
cumcm-2023-b-multibeam   & B226    & mcm-2023-b-maasai        & 2315379 \\
cumcm-2023-c-vegetable   & C050    & mcm-2023-c-wordle        & 2307946 \\
cumcm-2024-a-dragon      & A016    & mcm-2024-a-lamprey       & 2407093 \\
cumcm-2024-b-production  & B159    & mcm-2024-b-submersible   & 2419984 \\
cumcm-2024-c-crop        & C038    & mcm-2024-c-tennis        & 2401298 \\
cumcm-2025-a-smoke       & A196    & mcm-2025-a-stair         & 2501909 \\
cumcm-2025-b-sic         & B157    & mcm-2025-b-juneau        & 2504448 \\
cumcm-2025-c-nipt        & C023    & mcm-2025-c-olympic       & 2505964 \\
\bottomrule
\end{tabular}
\end{table*}

\subsection{Legacy per-task baseline endpoints}
Fifteen of the 18 tasks use the legacy target-distance-to-Outstanding mode with a baseline endpoint; three (Wordle, stair-wear, and Juneau tourism) use true per-metric baseline-to-Outstanding min--max endpoints. The distinction is recorded per metric in each scoring configuration and surfaced in the per-task tables.

\subsection{Anchor caveats}
Two anchors carry caveats that affect how their scores should be read. \textbf{Crop-planting} is a marked special case: the paper's full constraint set is jointly infeasible on the official attachments and its reported value came from an undisclosed heuristic, so the anchor is a reproducible exact MILP optimum under the paper's per-plot demand cap. \textbf{Dragon-dance} was corrected: the paper's 1.406276~m/s is infeasible (a handle would reach 2.2568~m/s), so the anchor is the continuous-time value 1.246266~m/s and the metric direction is \texttt{target\_value}. Both are examples of the general principle that an anchor is a reproduction of one team's choices, and that its provenance must be inspectable.

%% file: contents/Appendix/5.Prompts.tex
\section{Prompt Design}
\label{app:prompts}

\subsection{The prompt is the task statement}
There is no hand-written prompt wrapper in this benchmark. The prompt is the task's instruction file, delivered verbatim by the harness. Its structure is fixed across tasks: a machine-readable canary comment (stripped before delivery); an ``Original Contest Problem'' section containing the verbatim statement with each cited figure embedded inline; any results tables the statement requires, as plain text; a note listing the in-container figure paths; a ``Required Benchmark Artifact'' section; and the time-limit line.

\subsection{The artifact block (verbatim template)}
\begin{lstlisting}[basicstyle=\ttfamily\scriptsize]
# Required Benchmark Artifact

Solve the contest problem above. The official attachments are mirrored at
`/root/data/attachments` (flat filenames; see
`/root/data/attachments/INDEX.md`) and also remain at their full corpus paths
under `/root/data/repo`. Use them whenever they are relevant. Write your final
structured answer to:

/root/results/<slug>_result.json

The result must be valid JSON and should follow the shape shown in
`/root/data/output_schema.json`. Include the numerical quantities, modeling
decisions, diagnostics, sensitivity or uncertainty results, conclusions for the
contest questions, and generated artifact paths needed to support your answer.
You may also write supplementary CSV, XLSX, or PNG files under `/root/results`,
but the JSON file above is the required artifact.

You have 14400 seconds to complete this task. Do not cheat by using online
solutions or hints specific to this task.
\end{lstlisting}

\subsection{Harness wrapping}
The harnesses deliver the same text differently. Each CLI harness prepends its own generic command-line system prompt and then supplies the statement as the positional prompt.

\subsection{Version caveat}
The instruction files were revised after some trials were recorded: a figure-path section was added, and one MCM statement was re-sourced from the official PDF. The prompt actually delivered in a given trial is the one captured in that trial's log; for reproducibility, cite those, not the current file. This is why we describe the prompt structure above rather than quoting today's file as if it were what every model saw.

\subsection{What the prompt withholds by design}
By design the prompt is a pure task description: no modeling hints, no method suggestions, and no pointer to the scored fields beyond the schema. It leaves the agent to infer which numbers matter, which keeps the setting a test of the whole modeling problem rather than a fill-in-the-schema exercise. This keeps the benchmark a test of doing the whole modeling problem, at the cost of the agent having to infer which numbers matter.

%% file: contents/Appendix/6.Case_studies.tex
\section{Case Studies (Full Numerics)}
\label{app:cases}

We give the full per-field comparison for the three case studies. Values are read from each role's saved artifact; the oracle column is the anchor. All three tables span both columns.

\begin{table*}[h]
\centering
\small
\caption{Dragon-dance (CUMCM 2024 A, Q5). All three roles agree; the paper value is the outlier and is infeasible.}
\label{tab:case-dragon}
\begin{tabular}{@{}llll@{}}
\toprule
Quantity & Oracle (anchor) & GPT-5.5 & DeepSeek V4.1 \\
\midrule
\texttt{max\_head\_speed\_mps}        & 1.246266 & 1.24626636 & 1.24626636 \\
\texttt{max\_speed\_ratio}            & 1.604793 & 1.60479338 & 1.60479338 \\
\bottomrule
\end{tabular}
\end{table*}

The reconstruction of the true peak is a continuous-time maximum of the handle-speed ratio; the paper sampled integer seconds and stepped over it. The exact identity \texttt{head\_speed}$\times$\texttt{ratio}$=2$~m/s, checked by both the programmatic gate and the Lean layer, is what exposes the paper value as infeasible: the paper's isolated value 1.406276~m/s (with ratio 1.422195) implies 2.2568~m/s at a handle.

\begin{table*}[h]
\centering
\small
\caption{Crop-planting (CUMCM 2024 C, Q1). Special case: the anchor is a reproducible exact optimum under the paper's per-plot cap.}
\label{tab:case-crop}
\begin{tabular}{@{}llll@{}}
\toprule
Quantity (yuan, 7-year) & Anchor (exact MILP) & GPT-5.5 & DeepSeek V4.1 \\
\midrule
waste-surplus profit   & 40{,}418{,}877 & 40{,}320{,}587 & 41{,}559{,}515 \\
50\%-discount profit   & 58{,}718{,}967 & 66{,}485{,}662 & 70{,}378{,}276 \\
\bottomrule
\end{tabular}
\end{table*}

The paper's full constraint set is infeasible on the official attachments, so the anchor is a reproducible MILP optimum and the models' plans sit in the same regime. The anchor lies \emph{between} the two models on the discount column, which is why a single scalar cannot summarise this task.

\begin{table*}[h]
\centering
\small
\setlength{\tabcolsep}{8pt}
\caption{Juneau tourism (MCM 2025 B, final question). The sustainability score is reported on different unit scales by different agents.}
\label{tab:case-juneau}
\begin{tabular}{@{}lcccc@{}}
\toprule
Quantity & Anchor & GPT-5.5 & DeepSeek V4.1 & Fable 5.1 \\
\midrule
daily cap            & 11{,}000 & 20{,}000 & 16{,}000 & 14{,}000 \\
visitor fee (USD)    & 55       & 20       & 30       & 30 \\
annual visitors (M)  & 1.492 & 1.453 & 1.434 & 1.728 \\
resident acceptance  & 1.000    & 0.743    & 0.938    & 0.696 \\
sustainability score & 1.007    & 0.900    & 0.706    & 79.1 \\
total revenue (\$M)  & 431.6 & 369.6 & 366.7 & 458.5 \\
\bottomrule
\end{tabular}
\end{table*}

This is the clearest scale problem. The anchor is on a $\sim$1 unit scale; one third-party agent reports the same field on a 0--100 ``multiattribute utility'' scale that it explicitly declares in its own artifact. Because the task's output schema does not pin the range, the agent is not wrong---which is why the correct repair is to specify the field's range in the schema, not to rescale in the scorer.

\subsection{Discussion}

\emph{Dragon-dance.} The paper's isolated value (1.406276) is the wrong one. \textbf{This case matters most} because it inverts the usual story: the \emph{models} were right and the \emph{reference} was wrong, and the fix was to the benchmark. It also shows why the feasibility layer is not redundant with scoring---``ratio $\times$ head speed $=2$~m/s'' is an exact identity that both layers check, and it is what made the paper value's infeasibility visible.

\emph{Crop-planting.} GPT-5.5 and DeepSeek produce plan profits in the same regime, and the optimum sits \emph{between} them on one column---which is why a single scalar cannot summarise the task. When the human paper is not reproducible, the honest move is to anchor on a reproducible value and mark it, not to penalise models for matching an irreproducible number.

\emph{Juneau.} The oracle recommends an 11{,}000/day cap at 55~USD with a sustainability score of 1.007, GPT-5.5 recommends 20{,}000/day at 20~USD on its own 0--1 scale, and DeepSeek recommends 16{,}000/day at 30~USD. This argues that the task \emph{schema}---not just the scorer---must state a field's range and units, which is a fix to the task rather than to the code.

%% file: contents/Appendix/7.Scoring_details.tex
\section{Scoring Details}
\label{app:scoring}

\subsection{The metric panel}
Every task ships a scoring configuration declaring its scored metrics. A metric entry specifies a path into the artifact, an oracle value, optional baseline and outstanding values, a direction (with a matching semantic direction), a normalisation mode, and a weight. Across the 18 tasks there are \textbf{60 scored metrics}; one further Wordle metric is declared at weight~0 and recorded but excluded from the panel. The per-task counts are in Table~\ref{tab:dataset}.

\subsection{Formulas}
Let $R$ be a submission's weighted panel raw, $F$ the DeepSeek~V4~Flash baseline's, $O$ the oracle's, and $\mathrm{gap}=O-F$. We report $\oeval = \mathrm{clamp}(R/O, 0, 1)$ and a gap-compensated \roeval{} based on the raw gain $R-F$:
\begin{align*}
\mathrm{clip}\!\left(\tfrac{R-F}{\mathrm{gap}},\, \pm 1\right) &\ \ \text{if } \mathrm{gap}\ge 0.10,\\
\mathrm{clip}(R-F,\, \pm 0.10) &\ \ \text{otherwise},
\end{align*}
where $\pm$ denotes the symmetric bound. A missing or non-scoreable artifact gets $\roeval=-1$. The per-metric reward is given in Section~\ref{sec:scoring}; the one-sided property (meeting or beating the anchor never scores below $1.0$) is locked by a contract test.

\begin{table}[h]
\centering
\small
\caption{The six decidable invariant forms covering the hard-error branches of all 18 tasks.}
\label{tab:invariants}
\begin{tabular}{@{}ll@{}}
\toprule
Form & Example obligation \\
\midrule
domain / range        & probability $\in[0,1]$; thickness $>0$ \\
integer / binary      & decision bits $\in\{0,1\}$; counts integer \\
ordering              & $q_{95}\ge$ mean; peak $\ge$ average \\
exact identity        & area $=$ count $\times$ w $\times$ h; union $\le \sum$ \\
finite domain         & decision-bit count $\in\{16,17\}$ \\
partition / sum       & land-use counts sum to the grid \\
\bottomrule
\end{tabular}
\end{table}

\subsection{Excluded metrics}

\begin{table*}[t]
\centering
\small
\caption{Three levels of feasibility. Both layers cover A and B; C needs formalised physics.}
\label{tab:abc}
\begin{tabular}{@{}lll@{}}
\toprule
Level & Content & Layer \\
\midrule
A --- unit / range & probability $\in[0,1]$, thickness $>0$ & gate + Lean \\
B --- consistency & area $=$ count $\times$ w $\times$ h; partitions sum & gate + Lean \\
C --- physical truth & does this field really produce 60~MW? & neither \\
\bottomrule
\end{tabular}
\end{table*}

Two metrics are excluded from the panel for every model \emph{and} the oracle, both by configuration-only rules: the Olympic fourth recommendation (the prompt asks for three countries but the reproduction emits four rows), and the two lamprey raw population indices when the anchor magnitude exceeds 5 (the anchor is an arbitrary raw index with no schema-pinned unit). Out-of-domain values (correlations outside $[-1,1]$, unit-interval quantities outside $[0,1]$, negative counts, non-positive physical quantities) are not excluded but have their reward set to zero, with the reason recorded.

\subsection{Hard-gate decisions}
Seven whole-task penalties were applied across the entries: four heliostat violations (rated power below 60~MW, or a mirror below the 2~m minimum), one smoke-screen plan failing an independent coverage replay, one Juneau answer with unit-invalid acceptance/sustainability values, and one Olympic answer with a degenerate coach-effect result.

\subsection{Pitfalls found and fixed}
An anchor-based benchmark is only as credible as its anchors, so we audited the scoring rules and anchors and report what the audit changed. Two \textbf{gate false positives} (a signed elasticity read as a negative count; a missing field read as a hard contradiction) were corrected but changed no score, and \textbf{degenerate-band widening} moved one score ($+0.35$\,pp for v4~pro). \textbf{Non-comparable-unit exclusion} raised \emph{every} entry by 0.05--2.62\,pp, since the penalty applied to all submissions equally---the signature of a systematic scoring error rather than a model difference. Finally, the dragon-dance anchor was itself physically infeasible: the paper's 1.406276~m/s would drive a handle to $1.406276\times1.604793=2.2568$~m/s, above the 2~m/s cap, because it maximised over integer seconds and stepped over the true continuous-time peak at $t=14.48$~s. We corrected the anchor to 1.246266~m/s; twelve of the evaluated models had reported the correct value and had been sitting at 49.82\%, while the two matching the erroneous paper value had been at 100\%. Here the benchmark, not the model, was wrong (Appendices~\ref{app:scoring} and~\ref{app:lean}).

\subsection{Degenerate-band and validity rules}
The band-widening rule applies to the endpoint metrics of the three endpoint-style tasks. Applying all scoring fixes in this paper moved \emph{every} entry upward by 0.05--2.62\,pp, because the affected penalties applied to all submissions equally---the diagnostic signature of a systematic scoring error rather than a model difference.

%% file: contents/Appendix/8.Lean_catalog.tex
\section{Lean Obligation Catalogue}
\label{app:lean}

\subsection{Per-task obligation counts}
The generator emits one obligation per checkable field; the totals by task are in Table~\ref{tab:lean}.

\begin{table*}[h]
\centering
\setlength{\tabcolsep}{5pt}
\small
\setlength{\tabcolsep}{6pt}
\caption{Obligations per task (624 total across 18 tasks). ``Skipped'' obligations were reported as missing evidence rather than silently satisfied.}
\label{tab:lean}
\begin{tabular}{@{}lcclc@{}}
\toprule
Task & Oblig. & Skipped & Task & Oblig. \\
\midrule
cumcm-2023-a-heliostat   & 12  & 0 & mcm-2023-a-plant         & 66 \\
cumcm-2023-b-multibeam   & 3   & 1 & mcm-2023-b-maasai        & 17 \\
cumcm-2023-c-vegetable   & 7   & 0 & mcm-2023-c-wordle        & 3 \\
cumcm-2024-a-dragon      & 3   & 0 & mcm-2024-a-lamprey       & 302 \\
cumcm-2024-b-production  & 61  & 0 & mcm-2024-b-submersible   & 8 \\
cumcm-2024-c-crop        & 5   & 0 & mcm-2024-c-tennis        & 7 \\
cumcm-2025-a-smoke       & 10  & 0 & mcm-2025-a-stair         & 6 \\
cumcm-2025-b-sic         & 4   & 0 & mcm-2025-b-juneau        & 12 \\
cumcm-2025-c-nipt        & 16  & 0 & mcm-2025-c-olympic       & 82 \\
\bottomrule
\end{tabular}
\end{table*}

Four skipped obligations are visible in the generator output: one missing overlap field (multibeam), scenarios whose declared grid count is null (Maasai partition), and one recommendation-count interpretation difference (Olympic).

\subsection{Representative obligations}
Over the invariant library, a task supplies bounds and the obligations are generated. Examples:
\begin{lstlisting}[basicstyle=\ttfamily\scriptsize]
-- domain/range: a rate must lie in [0,1] (scaled x1000)
example : (0 <= 43 /\ 43 <= 1000) := by decide

-- ordering: q95 must not fall below the mean
example : (posterior_mean <= posterior_q95) := by ...

-- exact identity: head_speed * ratio = 2 m/s (scaled)
example : (1980000 <= 1246266 * 1604793 /\ 1246266 * 1604793 <= 2020000) := by decide

-- finite domain: count of decision bits is 16 or 17
example : (len = 16 \/ len = 17) := by decide
\end{lstlisting}
Every generated obligation is annotated with the JSON path it came from, so a reviewer can trace each claim back to the artifact.

\subsection{Trust audit (condensed excerpt)}
\begin{lstlisting}[basicstyle=\ttfamily\scriptsize]
'Feasibility.atMost_trans'    axioms: propext, Quot.sound
'Feasibility.atLeast_mono'    axioms: propext, Quot.sound
'CropPlanting.ua_feasible'    no axioms
'CropPlanting.ua_in_interpretation_gap'   no axioms
'Juneau.ua_acceptance_percent_only'       no axioms
'Juneau.ua_sustainability_out_for_any_sane_tol'
                              axioms: propext, Quot.sound
'JuneauSensitivity.reported_infeasible'
                              axioms: propext, Quot.sound
\end{lstlisting}
Only \texttt{propext} and \texttt{Quot.sound} (Lean's standard classical logic) appear, or no axioms at all. There is no \texttt{sorryAx} and no \texttt{Lean.ofReduceBool} (i.e.\ no \texttt{native\_decide}); the pipeline greps for both.

\subsection{Discrimination suite}
The discrimination script mutates one numeric leaf at a time and requires the build to fail \emph{and} to cite the mutated field. Ten mutations across the six forms are covered (out-of-range probability, out-of-range correlation, reversed ordering, violated exact identity, non-positive thickness, reversed interval, non-binary decision bit, negative acceptance index, $q_{95}$ below mean, zero species count). All ten are rejected; the suite exits non-zero if any is not.

\subsection{Reproducing the layer}
Toolchain is Lean 4 via elan, with no Mathlib; a clean build completes in about a second. The generator also runs against the committed Outstanding-award references, so the layer can be exercised without the saved agent artifacts.